\documentclass[10pt,letterpaper]{article}

\usepackage{net2brain}
\usepackage{pslatex}
\usepackage{apacite}
\usepackage{graphicx}

\title{Net2Brain: A Toolbox to compare artificial vision models \\ with  human brain responses}
 
\author{{\large \bf Domenic Bersch (s7407845@stud.uni-frankfurt.de)} \\
Department of Computer Science, Goethe Universität, Frankfurt am Main, Germany
\AND {\large \bf Kshitij Dwivedi (dwivedi@em.uni-frankfurt.de)} \\
Department of Computer Science, Goethe Universität, Frankfurt am Main, Germany
\AND {\large \bf Martina Vilas (martina.vilas@esi-frankfurt.de)} \\
Department of Computer Science, Goethe Universität, Frankfurt am Main, Germany\\
Ernst Struengmann Institute for Neuroscience, 60528 Frankfurt, Germany
\AND {\large \bf Radoslaw M. Cichy (rmcichy@zedat.fu-berlin.de)} \\
Department of Education and Psychology, Freie Universität Berlin, Berlin, Germany\\
Berlin School of Mind and Brain, Faculty of Philosophy\\
Bernstein Center for Computational Neuroscience Berlin, Berlin, Germany
  \AND {\large \bf Gemma Roig (roig@cs.uni-frankfurt.de)} \\
Department of Computer Science, Goethe Universität, Frankfurt am Main, Germany}

\begin{document}

\newpage
\maketitle

\begin{center}
\noindent\rule{\linewidth}{0.5pt}\\[0.25em]
\textbf{\Large Important notice}\\[0.5em]
\textbf{This version of the paper is outdated.}\\
The final, peer-reviewed version has been published in \emph{Frontiers in Neuroinformatics} (2025).\\
Please cite the version of record: \url{https://doi.org/10.3389/fninf.2025.1515873}\\[0.25em]
\noindent\rule{\linewidth}{0.5pt}
\end{center}

\section{Abstract}
{
\bf
We introduce Net2Brain, a graphical and command-line user interface toolbox for comparing the representational spaces of artificial deep neural networks (DNNs) and human brain recordings. While different toolboxes facilitate only single functionalities or only focus on a small subset of supervised image classification models, Net2Brain allows the extraction of activations of more than 600 DNNs trained to perform a diverse range of vision-related tasks (e.g semantic segmentation, depth estimation, action recognition, etc.), over both image and video datasets. The toolbox computes the representational dissimilarity matrices (RDMs) over those activations and compares them to brain recordings using representational similarity analysis (RSA), weighted RSA, both in specific ROIs and with searchlight search. In addition, it is possible to add a new data set of stimuli and brain recordings to the toolbox for evaluation. We demonstrate the functionality and advantages of Net2Brain with an example showcasing how it can be used to test hypotheses of cognitive computational neuroscience.

}
\begin{quote}
\small
\textbf{Keywords:} 
Toolbox; DNN; CNN; ViT; RSA; fMRI; MEG; Searchlight Analysis
\end{quote}

\section{Introduction}
Several studies have demonstrated the potential of DNNs to serve as state-of-the-art computational models of the primate visual cortex \cite{cadieu_deep_2014, CNN_Seyed,yamins_performance-optimized_2014,guclu_deep_2015,cichy_comparison_2016}. In the last decade, DNNs trained to perform visual tasks have successfully been able to resemble, predict and explain neural activity in the visual cortex. Different implementations of these models (varying, for example, their architecture, objective function, or training algorithm) have been compared to uncover the computational principles, algorithms and neurobiological mechanisms behind visual processing \cite{FrameworkNeuro}.

To promote this line of research, new benchmarks, datasets, and challenges relevant to cognitive neuroscience experiments have been developed \cite{Algonauts,cichy2019algonauts,cichy2021algonauts,BrainScore1}. However, to fully take advantage of these models and frameworks, a toolbox for efficiently comparing the representational spaces of state-of-the-art DNNs and brain responses is needed. Some toolboxes have been developed to facilitate the use of DNNs, however, they tend to focus only on a small subset of supervised image classification models, even though studies have shown that DNNs trained for different tasks can also help to provide new information about the visual cortex \cite{tang2021cortical,DwivediPLOSCOMPBIOL21}.

We, therefore, introduce Net2Brain, an easy-to-use toolbox that allows neuroscientists to efficiently incorporate over 600 DNN trained for different objective functions, datasets, etc, into their research. We opensource it to promote its continual growth over time.

\section{Related Work}
In the past, deep learning models have been adopted across scientific fields to answer domain-specific questions \cite{raghu2020survey}. This was greatly facilitated by open-source software that allows the straightforward usage and development of DNNs, such as PyTorch \cite{pytorch}, Tensorflow~\cite{tensorflow}, Caffe~\cite{caffe} and Keras~\cite{keras}.
With such a variety of libraries at hand and the increasing use of deep learning models in neuroscience research, recent toolboxes have been developed to facilitate synergy between both fields.
The rsatoolbox \cite{nili2014toolbox} provides functions for comparing the representational space of computational models and brain responses. This software library expects the user to provide as input the already extracted activations of a DNN.
BrainScore \cite{BrainScore1,BrainScore2} and THINGSvision \cite{thingsvision} are toolboxes that extend this functionality and allow computing feature representations from some DNNs as well as compare them with brain recordings.
However, these libraries implement DNNs that were mainly developed for image classification tasks.
This sub-selection limits the use of this approach when examining the neural representations of humans performing other perceptual and cognitive functions.
Net2Brain expands the DNNs available for comparison from supervised models trained on image classification, instance and panoptic segmentation, 3D scene understanding, and action recognition tasks, to self-supervised models \cite{caron2020unsupervised,he2020momentum} and multimodal DNNs \cite{radford2021learning}. We further recognize the importance of video datasets which could provide new insights into the human processing of motion and event understanding. %Net2Brain can extract DNNs' responses to this type of input modality.

\section{Net2Brain}
Net2Brain is based on the ideas and goals of the Algonauts project \cite{Algonauts}. This intuitive toolbox provides all the functionality needed for rapidly extracting the representations of a variety of DNNs, computing their representational dissimilarity matrices (RDMs), and comparing them to brain datasets. 
It employs RSA, weighted RSA, to make this comparison, and provides an in-depth examination of the correlation between the representational space of brain datasets and DNNs, for specific ROIs or in searchlight fashion.
In addition, Net2Brain also informs about the quality of the brain recording being inspected and provides the flexibility to add new datasets and DNNs for analysis. 
Users can test a new hypothesis with a few clicks via CLI-Commands, a command-line interface ideal for servers like Google Colab, or a conveniently-designed GUI.

Each of the over 600 implemented neural networks in Net2Brain is capable of processing image and video input data (.jpg and .mp4), providing the opportunity of studying cognitive functions associated with the processing of continuous stimuli. In addition, the toolbox introduces models trained on a variety of visual tasks. Although image classification Convolutional Neural Networks (CNNs) have shown the best predictive power of the visual cortex \cite{yamins_performance-optimized_2014,CNN_Seyed}, models trained on other tasks could improve our understanding of the neural processing of a wider range of perceptual and cognitive functions. Hence we included models trained for scene understanding \cite{taskonomy_DBLP:journals/corr/abs-1804-08328}, segmentation models,  \cite{wu2019detectron2}, video models \cite{fan2020pyslowfast}, multimodal models \cite{radford2021learning}, and self-supervised models\cite{caron2021emerging}.

Using these models, features can be generated to be compared with available brain datasets. The evaluation function of Net2Brain allows the simultaneous comparison of the RDMs of multiple DNNs and brain datasets using RSA and weighted RSA. As an output of this step, the toolbox supplies a graph with the squared correlation coefficient per layer obtained through the analysis, along with a measure of statistical significance, and an estimate of the lower and upper noise ceiling of the brain responses. The computed data and the resulting graph are automatically stored in the filesystem to be easily accessed.
The toolbox can be downloaded from GitHub (https://github.com/ToastyDom/Net2Brain.git) and also contains the fMRI and MEG datasets used in the 2019 Algonauts challenge \cite{Algonauts}, provided in RDM format. Providing these datasets enables the user to immediately test the functionality of the program, and intuitively shows how to add new brain recordings to the toolbox.

\section{Prediction of brain responses using multimodal DNNs}

In the last few years, the field of deep learning has shown that DNNs trained on multi-sensory input, which are capable of creating multimodal representations, achieve better generalization and overall performance. In this context, much debate exists in the field of cognitive neuroscience on the multimodal nature of cortical representations, and the idea that brain areas higher up in the hierarchy might need to encode these types of representations for carrying out more abstract computations \cite{tang2021cortical}. Combining both fields, this hypothesis could be tested by analyzing if brain representation are more similar to multimodal DNNs than unimodal ones.

As an exploratory work, we used Net2Brain to compare the responses of the multimodal CLIP-ResNet50 and CLIP-ViT-B/32, a self-supervised DNN trained on image-text pairs~\cite{radford2021learning}, with its unimodal counterparts ResNet50 and ViT-B/32, which are supervised DNN trained to perform object recognition on Imagenet, to human functional magnetic resonance imaging (fMRI) recordings from the dataset by Michael F. Bonner et al. \cite{Bonner201618228}.

\begin{figure}[t!]
    \centering
    \includegraphics[width=\linewidth]{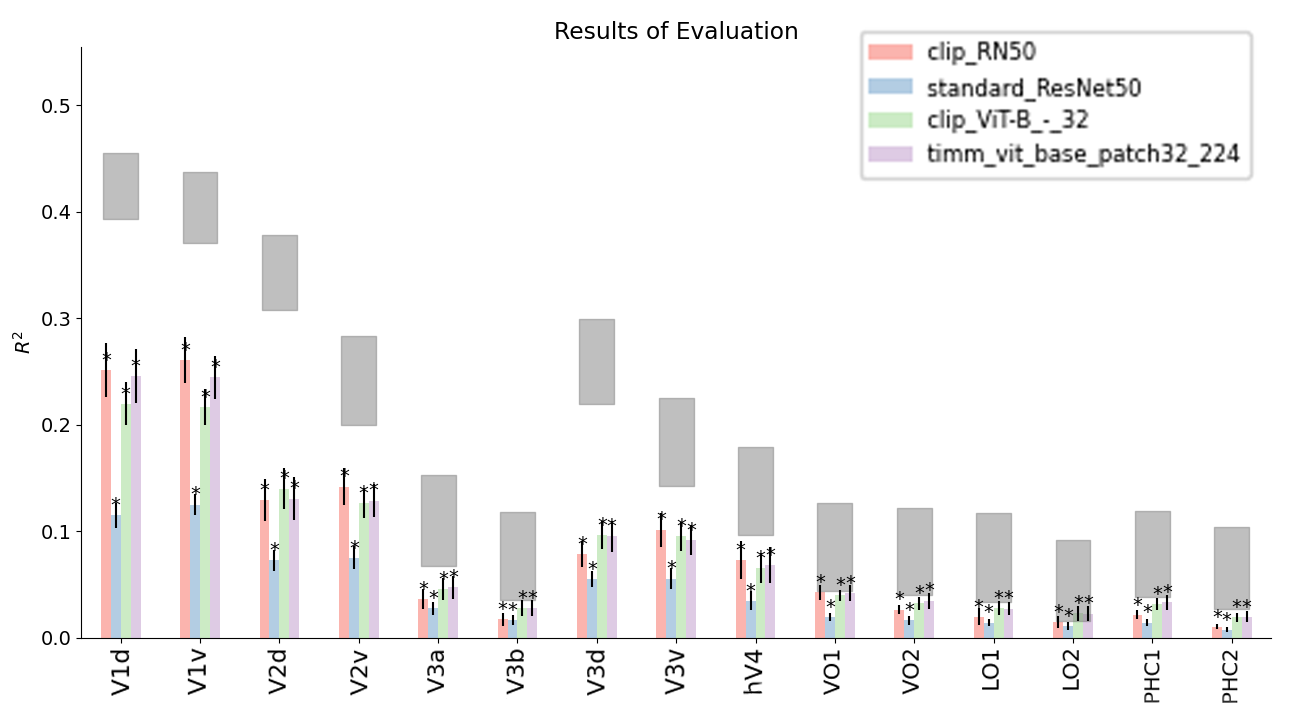}
    \includegraphics[width=\linewidth]{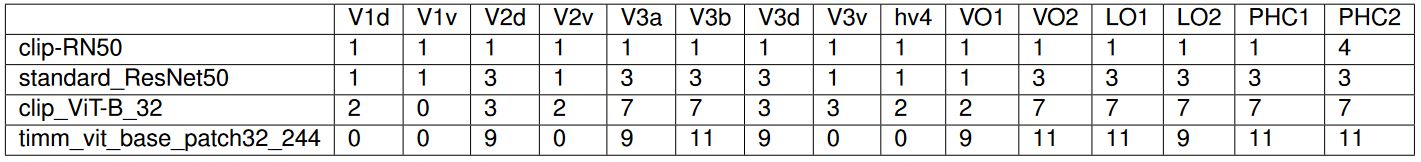}
    \caption{Prediction of brain responses using multimodal DNNs vs their unimodal counterparts in the ROIs in Fig.~\ref{fig:brain} and a table displaying the layers with the highest correlation. The range from lower to upper noise ceiling is indicated by the gray box and the asterisk above the bars indicates the significance of the calculated data. The error bar represents the standard error across subjects.\vspace*{-0.3cm}}
    \label{fig:plot}
\end{figure}

\begin{figure}[t!]
    \centering
    \includegraphics[width=1\linewidth]{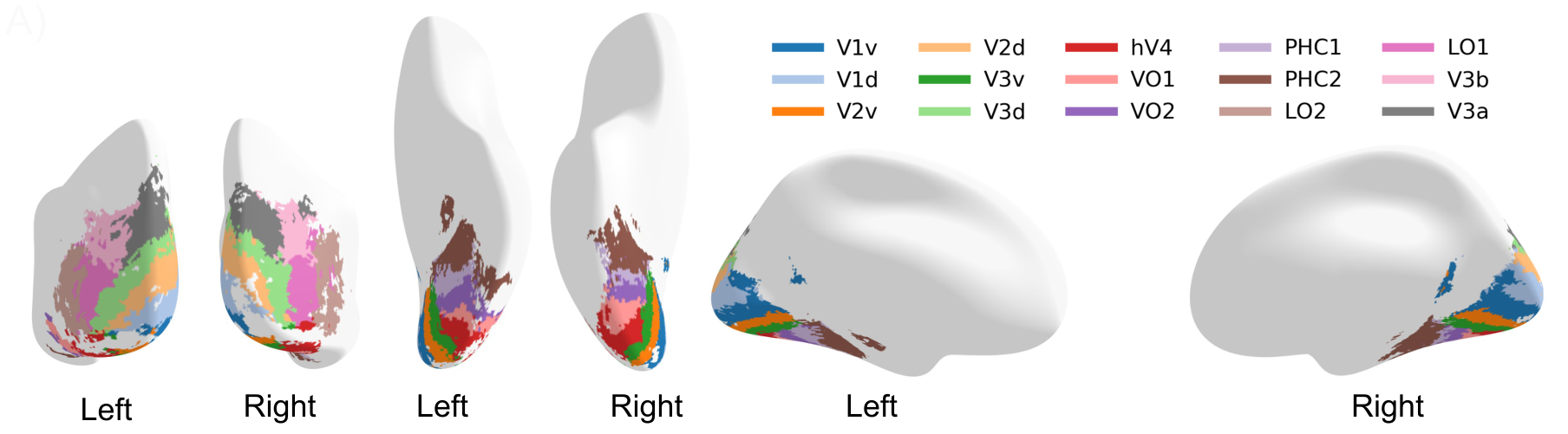}
    \caption{Cortical overlay showing locations of cortical regions from the probabilistic atlas used in Fig.~\ref{fig:plot}.\vspace*{-0.3cm}}
    \label{fig:brain}
\end{figure}

As illustrated in Fig.\ref{fig:plot}, we found that the multimodal CLIP-ResNet50 has significantly better predictability of the regions of interest (ROIs), which are displayed in Fig.~\ref{fig:brain}, than its unimodal counterpart ResNet50 throughout all presented layers. This can be seen as a prelude toward research that argues whether the inclusion of captions allows encoding spatial relations and how other modalities could improve predictability.

Another pattern that can be observed is that although CLIP-ViT and normal ViT behave similarly, they both have better predictability of the regions than ResNet50. This invites to delve deeper into exploring regions of the brain using other DNNs rather than CNNs, and having different architectures to help understand the structure of the visual cortex.

In sum, Net2Brain facilitates investigating correlations between different DNNs and brain ROIs and reveals exciting patterns that can be further explored.

\section{Conclusion}
We have introduced Net2Brain, a toolbox for comparing the responses of artificial neural networks and the human visual cortex using representation similarity analysis. Our toolbox facilitates the adoption of DNNs in cognitive neuroscience research, lowers the knowledge barrier for newcomers that want to implement these tools, and provides users the flexibility to carry out these analyses using their computational models and brain datasets. We have also demonstrated the simplicity of using Net2Brain for testing a hypothesis from cognitive computational neuroscience. In the future, the toolbox will include more brain datasets and functions for carrying out common analyses in neuroscience research, such as variance partitioning analysis and encoding models.

\section{Acknowledgments}
This work was funded with the support from the Alfons and Gertrud Kassel Foundation (G.R.), by the German Research Foundation (DFG, CI241/1-1, CI241/3-1 to R.M.C.) and by the European Research Council (ERC, 803370 to R.M.C.).

\bibliographystyle{apacite}

\setlength{\bibleftmargin}{.125in}
\setlength{\bibindent}{-\bibleftmargin}

\bibliography{net2brain}
\end{document}